\documentclass[12pt]{article}

\usepackage{newtxtext,newtxmath}

\usepackage{graphicx}

\usepackage[letterpaper,margin=1in]{geometry}

\renewenvironment{abstract}
	{\quotation}
	{\endquotation}

\date{}

\makeatletter
\renewcommand{\fnum@figure}{\textbf{Figure \thefigure}}
\renewcommand{\fnum@table}{\textbf{Table \thetable}}
\makeatother

\usepackage{scicite}

\usepackage{url}

\def\scititle{
	A Robotic Testing Platform for Pipelined Discovery of Resilient Soft Actuators
}
\title{\bfseries \boldmath \scititle}

\author{
	Ang~(Leo)~Li$^{1,2\ast}$,
	Alexander~Yin$^{3}$,
	Alexander~White$^{3}$, 
    Sahib~Sandhu$^{3}$,\and
    Matthew~Francoeur$^{4}$,
    Victor~Jimenez-Santiago$^{3}$,
    Van~Remenar$^{3}$,\and
    Codrin Tugui$^{5}$,
    Mihai Duduta$^{1,2,3\ast}$ \and
	\small$^{1}$Department of Mechanical and Industrial Engineering, University of Toronto, \and \small Toronto, ON, M5S 3G8 Canada\and
	\small$^{2}$Institute of Materials Science, University of Connecticut, Storrs, CT, 06269 U.S.A.\and
    \small$^{3}$School of Mechanical, Aerospace, and Manufacturing Engineering,\and
    \small University of Connecticut, Storrs, CT, 06269 U.S.A.\and
    \small$^{4}$Material Science and Engineering, \and 
    \small
    University of Connecticut, Storrs, CT, 06269 U.S.A.\and
    \small$^{5}$Inorganic Polymers Department, Petru Poni Institute of Macromolecular Chemistry, \and \small Iasi, 700487 Romania\and
	\small$^\ast$Corresponding author. Email: shffleo.li@mail.utoronto.ca, mihai.duduta@uconn.edu
}

\begin{document} 

\maketitle

\newpage
\begin{abstract} \bfseries \boldmath
Short lifetime under high electrical fields hinders the widespread robotic application of linear dielectric elastomer actuators (DEAs). Systematic scanning is difficult due to time-consuming per-sample testing and the high-dimensional parameter space affecting performance. To address this, we propose an optimization pipeline enabled by a novel testing robot capable of scanning DEA lifetime. The robot integrates electro-mechanical property measurement, programmable voltage input, and multi-channel testing capacity. Using it, we scanned the lifetime of Elastosil-based linear actuators across parameters including input voltage magnitude, frequency, electrode material concentration, and electrical connection filler. The optimal parameter combinations improved operational lifetime under boundary operating conditions by up to 100\% and were subsequently scaled up to achieve higher force and displacement output. The final product demonstrated resilience on a modular, scalable quadruped walking robot with payload carrying capacity ($>$100\% of its untethered body weight, and $>$700\% of combined actuator weight). This work is the first to introduce a self-driving lab approach into robotic actuator design. 
\end{abstract}

\noindent
Reduced lifetime under high electrical fields limits the practical use of linear dielectric elastomer actuators (DEAs) in aerial \cite{chen2019controlled, shintake2014foldable}, aquatic \cite{berlinger2018modular, flores2025robonautilus}, walking \cite{ji2019autonomous, pei2003multifunctional}, and inspection \cite{tang2022pipeline, sandhu2025miniature} robots. While most actuator designs achieve long lifetimes at low electrical fields, their durability drops sharply under higher fields. Yet, maximum power, strain, and force outputs are realized only near high-field operation (80\% of the breakdown field) due to the near-exponential dependence of mechanical output on electric field strength \cite{hajiesmaili2021dielectric}. Although the community commonly reports DEA longevity in terms of "cycle life", we find that "lifetime" is a more meaningful metric, since high-frequency actuation can yield large cycle numbers that do not accurately reflect material degradation \cite{jiang2022long}. Despite its critical importance, DEA lifetime research is challenging: obtaining such data is time-intensive, requiring prolonged per-sample tests that occupy valuable instruments such as high-voltage power supplies, oscilloscopes, function generators, and force or displacement sensors. Furthermore, lifetime spans a multidimensional parameter space, including voltage, frequency, material composition, and processing conditions. The high resource demand and complexity make systematic lifetime mapping impractical using conventional manual testing approaches.

Developing stand-alone, high-throughput, and modular testing instruments is fundamental to advancing DEA lifetime research. Previous efforts have focused on building multi-channel testing setups \cite{kuhnel2021automated,matysek2011lifetime}, laying the groundwork for studying the evolution of electrical properties throughout operation \cite{li2022real,jiang2022long,li2024data} and the influence of environmental conditions\cite{albuquerque2021influence}. The next step is to treat DEA lifetime as a multidimensional space, uncovering how voltage, frequency, and material choices affect longevity. Advances in self-driving labs (SDLs) have demonstrated the power of robotic experimental platforms and algorithms to automate data-intensive, high-dimensional, and time-consuming research\cite{tabor2018accelerating}, ranging from optimizing digital composites\cite{li2024computational} to high-throughput semiconductor material discovery\cite{macleod2020self} and characterization\cite{siemenn2025self}. Partial SDL implementations have also advanced battery cycle-life prediction using machine learning algorithms on collected dataset\cite{severson2019data}, and honeybee colony monitoring through autonomous robotic data collection\cite{ulrich2024autonomous}. Inspired by these developments, we adopt the SDL approach to introduce robotic testing infrastructure into DEA cycle-life research, aiming to reduce manual effort while expanding testing throughput and parameter-space coverage.

In this work, we propose a optimization pipeline for geometrically linear DEAs that spans from parameter scanning to downstream integration of optimized actuators into mobile robots. The pipeline described in \textbf{Figure 1} is enabled by a novel testing robot that integrates multi-channel programmable voltage input and electromechanical measurement capabilities, including current, voltage, impedance, displacement, and blocked force. A stepper-motor-based automated system switches actuators between force and displacement measurements. Batches of simplified linear DEAs (10 active layers, 1.2 cm tall) made of Elastosil P7670 \cite{ren2022high} elastomer are fabricated and tested using this system. The lifetime scanning procedure starts from the best-known material configuration, samples are tested under a DC square wave to scan frequency and electric field, identifying the optimal operating range that yields the longest lifetime while maintaining strong mechanical performance.

\begin{figure*}[!ht]
\centering
\includegraphics[width=\linewidth]{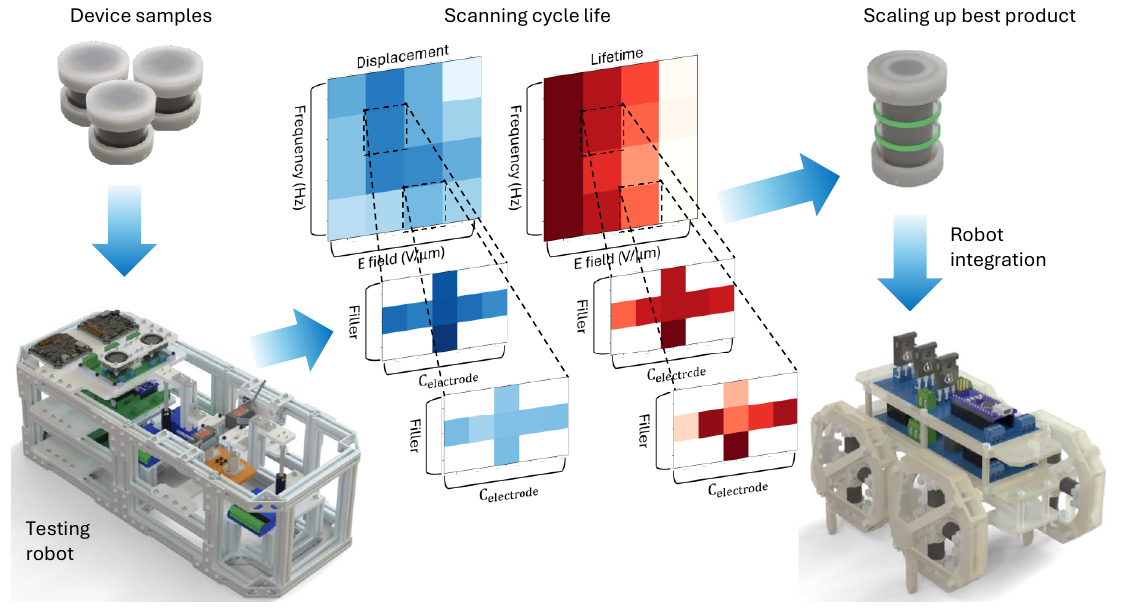}
\caption{\textbf{Linear DEA optimization pipeline.} Batches of linear DEA samples are fabricated and tested using the DEA testing robot. lifetime is scanned across the parameter space defined by voltage, frequency, and material choices. The optimal parameter combination is then used to scale up the actuator in size and layer number, achieving higher displacement and force output. Finally, the optimized actuators are integrated into a quadruped walking robot.}
\label{fig:f1}
\end{figure*}

In the next stage, actuators fabricated with varying electrode concentrations and electrical connection filler materials are tested at the identified optimal voltage field and frequency to determine the best material combinations. The structure of linear actuators and their operating principle is shown in \textbf{Figure S1}. Electrical connections, often overlooked in DEA research, play a critical role in device durability. We found that the conductive filler used at the electrical connection can significantly influence lifetime performance (up to 500\% among materials scanned in this work). The optimal parameters identified through this pipeline are then used to fabricate scaled-up actuators with increased layers (20 active layers), size (2 cm tall), and added reinforcement to enhance axial strain. These optimized actuators are integrated into a modular, scalable quadruped robot. The robot demonstrates the payload capacity of $>$100\% of its tethered body weight and $>$700\% of combined actuator weight. The demonstration highlights not only the actuator’s resilience but also the potential of non-metallic, non-magnetic, compliant electrical actuators for quadruped locomotion in extreme environments such as strong electromagnetic fields.

This work demonstrates a systematic and automated approach to investigating one of the most critical and resource-intensive attributes of DEAs: operational lifetime. It is also the first to bridge the self-driving lab (SDL) concept with robotic actuator optimization. The testing robot, built from open-source electronics and off-the-shelf instruments, can be easily scaled beyond two channels and serves as a data-collection engine for DEAs. The collected data can potentially support model training for performance prediction, rapid evaluation, and material design. Future work will integrate automated sample fabrication and data-driven optimization to evolve the system into an autonomous DEA design platform. Our vision towards a closed-loop, device-level SDL, using DEAs as a case example, is outlined in the discussion section. 

\subsection*{Results}
\subsubsection*{Testing robot}
The actuator testing robot is the central enabling component of the discovery pipeline. Its design principle is to integrate a comprehensive set of electromechanical measurement capabilities into a standalone system \textbf{(Figure 2A)}. The robot has a benchtop-compatible physical profile, allowing easy integration into existing lab setups. Open-source electronics and measurement instruments enhance system integration at the programming level. Key hardware components, including high-voltage converters, laser displacement sensors (LDS), and force sensors, are mounted in a plug-and-play configuration, enabling quick replacement or upgrades to accommodate different power and measurement ranges.

The proposed robot has a two-channel setup, where each channel has an independent power supply and measurement track, but both are controlled by a single controller set \textbf{(Figure 2B)}. This architecture balances scalability with integrated data collection and control. For mechanical property measurement, each channel includes an LDS for displacement measurement and a force sensor for blocked-force testing. Since displacement and blocked force cannot be measured simultaneously, a motorized switching mechanism is integrated to alternate between the two modes. A rotary stepper motor carries a plate that holds the linear DEA samples for each channel. By rotating, it positions the actuators beneath either the LDS or the force sensor. Blocked-force measurement requires restricting actuator motion; to achieve this, the force sensor is mounted on a linear motor that automates constraining and releasing. During testing, the linear motor lowers the force sensor onto the DEA sample and uses the force reading as feedback until a predefined threshold is reached, after which the actuator is released \textbf{(Movie S1)}.

\begin{figure*}[!ht]
\centering
\includegraphics[width=\linewidth]{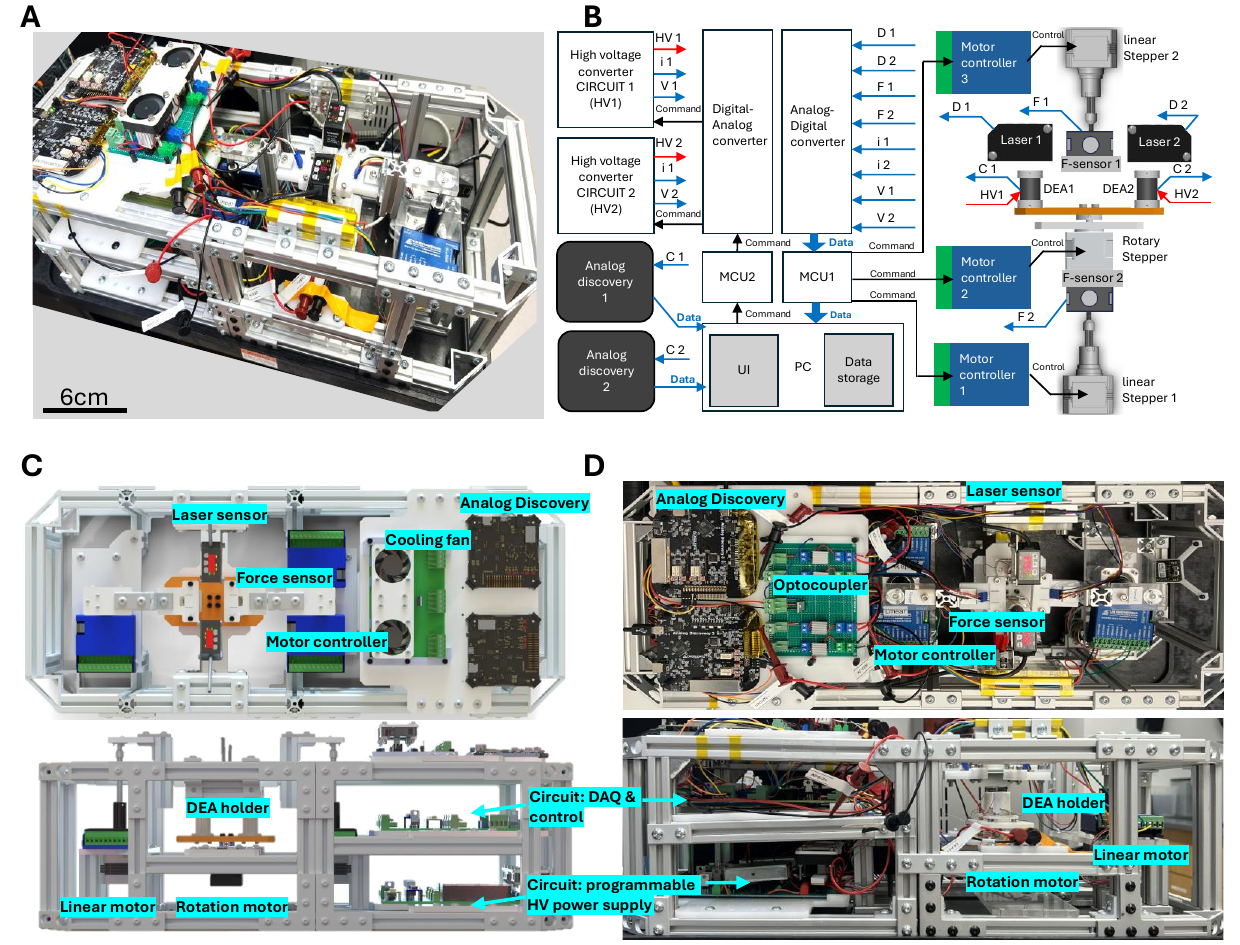}
\caption{\textbf{The linear DEA testing robot.} \textbf{A,} the testing robot. \textbf{B,} the block diagram of the system architecture. \textbf{C,} the top-down and side views of the testing robot. \textbf{D,} the photos of the top-down and side views of the testing robot.}
\label{fig:f2}
\end{figure*}

The main electronics are organized into two stacked circuit boards: one for data collection and control, and another as a programmable high-voltage power supply \textbf{(Figure 2C and D)}. The top board integrates the analog–digital converter (ADC) interface for all electromechanical measurements, including displacement, force, and current–voltage feedback from the power supply \textbf{(circuit diagram, Figure S2)}. It also provides motor control interfaces. A microcontroller (MCU: Arduino Nano) handles data acquisition and automatic motor control. The same MCU controls the high-voltage opto-couplers (HVM OR-100), which isolate the DEAs and the LCR meters (embedded in the top-layer circuit) from the power supply when switching measurement modes. The bottom board serves as a dual-channel programmable power supply \textbf{(circuit diagram, Figure S3)}. Each channel independently controls voltage magnitude, frequency, and waveform. Default high-voltage converters are XP Power HRC series (5 W). A user interface (UI) programmed in Python enables communication between the Arduinos and the PC. The UI connects to both Arduino MCUs and manages data acquisition. When the DEAs are isolated from high voltage, the robot performs impedance measurements using two Digital Analog 3 (DA3) units. These DA3s are controlled via manufacturer-provided APIs integrated into the UI. The overall architecture and workflow of the testing robot is shown in \textbf{Figure S4}.

\subsubsection*{DEA lifetime optimization pipeline}
We developed and implemented a staged lifetime optimization approach that first scans the boundary conditions of the operating regime, including electric field and frequency, followed by material parameters to extend lifetime under these boundary conditions. Here we define the operational lifetime as the time before device displacement degrades to 80\% \cite{han2014comparative} of its initial value. This definition is inspired by practices in the battery industry and is more practically relevant than the 50\% criterion \cite{carpi2015standards}. The DEA samples are fabricated by rolling a multilayer DEA sheet and encapsulating it between two rigid caps for electrical connection \textbf{(Figure S1)}. Details of the multilayer fabrication process are provided in the Materials and Methods. The material parameters explored in this work include (i) the electrical connection filler and (ii) the carbon nanotube (CNT) electrode concentration. The electrical filler is the conductive material that interfaces the DEA with the electrical wire to form a reliable conductive path \textbf{(Figure S1)}. This filler is typically a colloidal material applied by brushing or painting. Through experimental practice, we found that the choice of filler material has a significant impact on device lifetime, yet this factor has received little attention in the DEA literature.

\begin{figure*}[!ht]
\centering
\includegraphics[width=\linewidth]{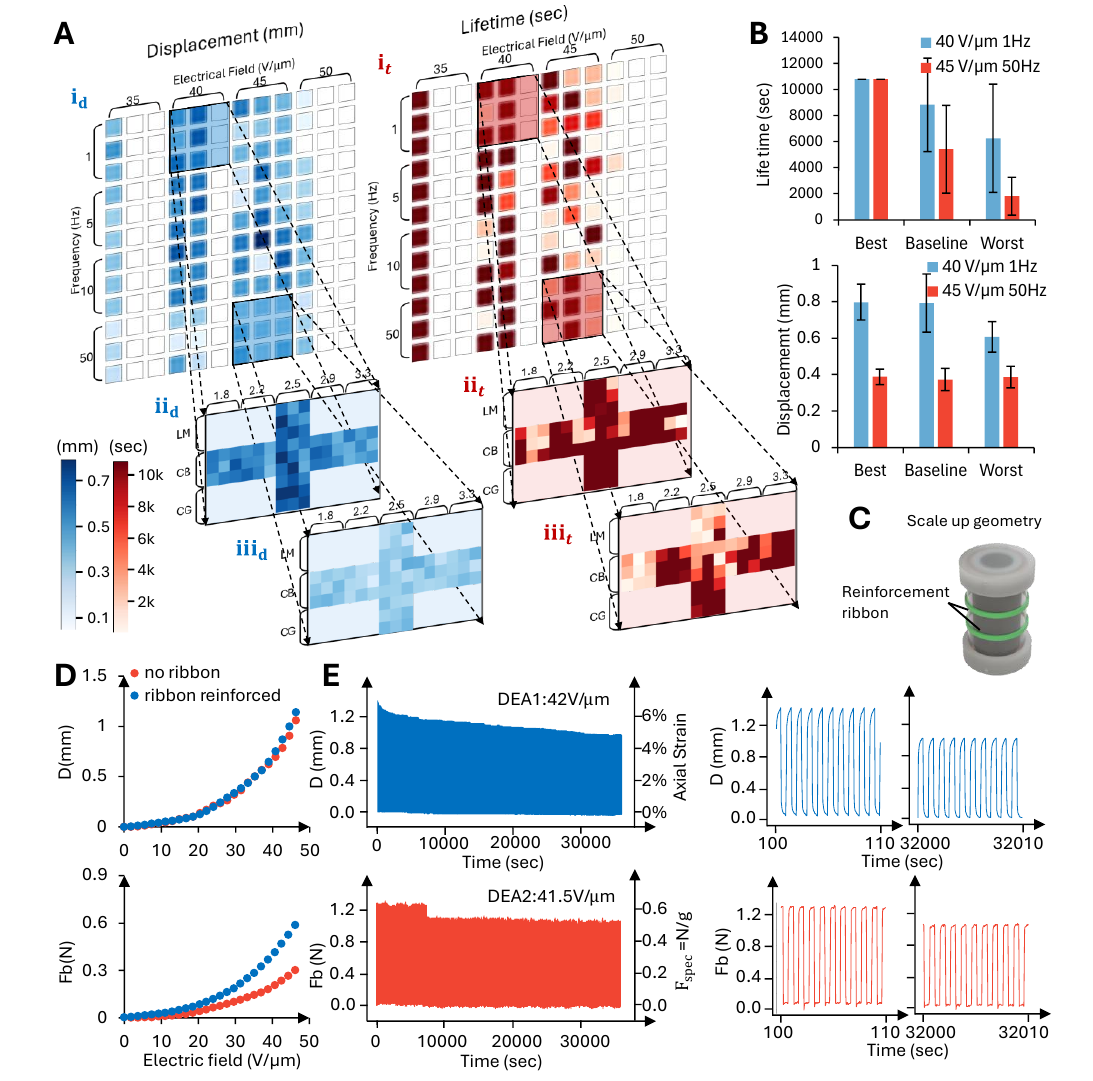}
\caption{\textbf{Parameter optimization for optimal lifetime and mechanical output.} \textbf{A,} layered optimization structure. \textbf{B,} Optimization result comparison. Lifetime and average displacement at 1Hz ($40,V/\mu m$)  and 50Hz ($45,V/\mu m$) of best, baseline (default) and worst material combinations are presented. \textbf{C,} 3D model of the scaled up DEA with reinforcement ribbons. \textbf{D,} comparison of displacement under the same electric field for scaled-up DEAs with and without reinforcement ribbons. \textbf{E,} displacement and force degradation of scaled-up DEAs at 1Hz, above $40,V/\mu m$, for 10 hours of continuous actuation.}
\label{fig:f3}
\end{figure*}

The first stage scans DEA lifetime across operating electric fields and actuation frequencies. The goal is to identify boundary conditions where displacement is high but lifetime is short, making improvement most impactful. At this stage, material choices starts with our best in-house practices: a carbon black–PDMS mixture is used as the conductive filler, and P3 CNT ink is used as the electrode at a concentration of 2.5 mL/FA (filtration area $FA = 50.24,\mathrm{cm}^2$). Batches of samples are fabricated and tested under DC square-wave actuation across electric fields of 35, 40, 45, and 50 V/$\mu$m and frequencies of 1, 5, 10, and 50 Hz. Lifetime (capped at 3 hours) and average displacement are used as evaluation benchmarks \textbf{(Figure 3A $(i_d)$ and $(i_t)$)}. At the lower electric field of 35 V/$\mu$m, device performance is stable, with all samples maintaining effective lifetimes beyond 3 hours, although the average displacement remains small. Experiments at this field were stopped after three samples per field–frequency pair due to low variation. At the highest field of 50 V/$\mu$m, devices consistently failed early (average lifetime $<$ 1500 s), indicating that this field is outside the practical operating range and too close to electrical breakdown. From these scans, we identified the boundary conditions to be 40 V/$\mu$m at 1 Hz and 45 V/$\mu$m at 50 Hz. At these conditions, devices exhibit the largest average displacement while still having potential for lifetime improvement.

The next stage focuses on the boundary conditions and scans two material parameters: the electrical connection filler and the CNT electrode concentration. Three filler materials are selected: liquid metal (LM), carbon black–PDMS mixture (CB, default), and carbon grease (CG). Five CNT concentrations are tested (1.8, 2.2, 2.5, 2.9, and 3.3 mL/FA). To maximize sample efficiency, a controlled-variable approach is used. When scanning filler materials, the CNT concentration is held constant at 2.5 mL/FA, and when scanning CNT concentration, the filler is fixed as CB. Average displacement and lifetime results are shown in Figure 3A $ii_d$ and $iii_d$. At 40 V/$\mu$m and 1 Hz, the best-performing combination is the having CG as the electrical connection filler and 2.5 mL/FA CNT concentration (default). At 45 V/$\mu$m and 50 Hz, the highest average displacement is achieved with a CNT concentration of 2.9 mL/FA while using CB as the filler. However, the longest lifetime under the same conditions is obtained using CG as the filler with the default CNT concentration of 2.5 mL/FA. Since the optimal parameters for displacement and lifetime were not tested together, the final stage fabricates devices using the combined parameters of 2.9 mL/FA CNT concentration and CG as the connection filler and tested them at 45 V/$\mu$m and 50 Hz. The results show improved performance in both displacement and lifetime.

Through the staged lifetime optimization approach, the best material combinations achieved a significant increase in lifetime while also improving the average displacement incrementally compared to the default case. \textbf{Figure 3B} summarizes these comparisons. At 1 Hz and 40 V/$\mu$m, the optimal configuration uses carbon grease as electrical connection filler and CNT concentration of 2.5 mL/FA. Under these conditions, the average lifetime increases by 22\% relative to the default baseline and by 72\% compared to the worst-performing case (1.8 mL/FA CNT with CB filler). At 50 Hz and 45 V/$\mu$m, the best combination uses carbon grease as the filler and 2.9 mL/FA CNT concentration, resulting in a 99\% lifetime increase over the baseline and a 496\% increase compared to the worst-performing configuration. In addition to lifetime gains, the optimized materials greatly reduce lifetime variation across samples, indicating more consistent device performance and better suitability for practical use.

We scaled up the optimal material combinations by increasing device length and the number of active layers to enhance force and displacement output for robotic applications. The actuator was enlarged by doubling the active layers from ten to twenty and increasing its length from 1 cm to 2 cm. Two reinforcement ribbons cut from heat-shrink tubing were wrapped around the rolled DEAs for strain limiting to improve mechanical performance \textbf{(Figure 3C)}. The effect of ribbon reinforcement is shown in \textbf{Figure 3D}. Displacement and maximum blocked force were measured across electric fields. The reinforcement does not affect displacement but doubles the maximum blocked force (photographic comparison in \textbf{Figure S5A}. 10 hours of continuous operation above 40 V/$\mu$m at 1Hz shows consistent performance with minimal degradation \textbf{(Figure 3E)}, maintaining an average axial strain of 5.5\% and a specific blocked force of 0.55 N/g. Frequency sweep of the actuator is performed and displacement is measured (\textbf{Figure S5B}).

\subsubsection*{Quadruped robot powered by linear DEAs}
A quadruped robot is demonstrated using the scaled-up linear DEAs. Prior to this work, even tethered DEA-based walking robots had not demonstrated meaningful payload capacity \cite{pei2004multiple, nguyen2018development, ji2019autonomous}. Robots using DEAs as artificial muscles to actuate rigid leg frames have also not achieved practical, self-supporting locomotion \cite{nguyen2014small}. The modular walking robot developed here consists of base locomotion units that can be assembled through simple connectors to form larger systems \textbf{(Figure 4A)}. When multiple base units are combined, each operates in the same sequence; therefore, regardless of the total number of units, only three high-voltage switching channels are required. This scalability greatly simplifies the electronics design for more complex robot control.

\begin{figure*}[!ht]
\centering
\includegraphics[width=\linewidth]{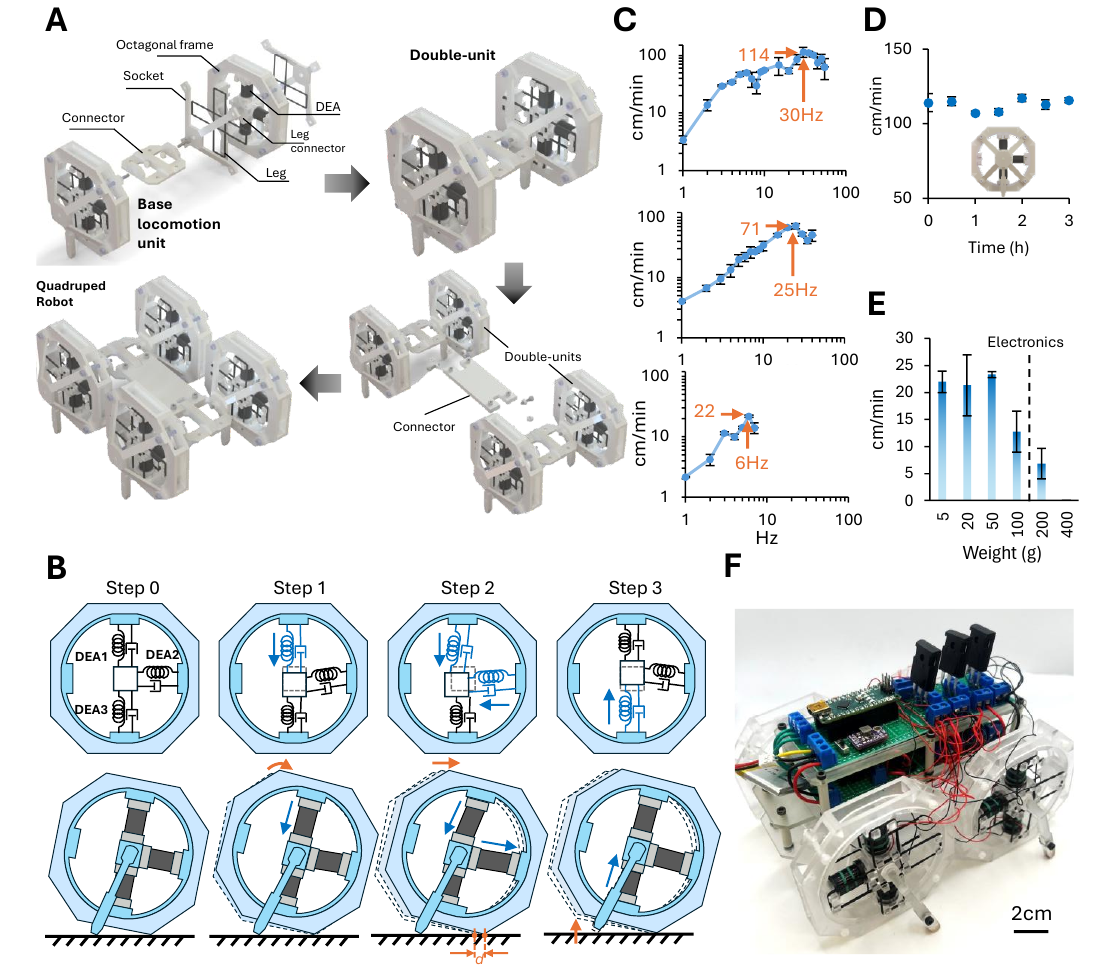}
\caption{\textbf{The Modular quadruped walking robot.} \textbf{A,} Robot design and assembly. Base locomotion units serve as the building blocks; two units form a double-unit module, and two double-units assemble into a quadruped robot. \textbf{B,} Gait mechanism of a single locomotion unit. \textbf{C,} Frequency response of walking speed for 1-, 2-, and 4-unit robot configurations. \textbf{D,} Long-term speed stability of a base locomotion unit. \textbf{E,} Payload-dependent walking speed of the tethered quadruped robot. \textbf{F,} Quadruped robot carrying onboard power electronics designed for untethered operation.}
\label{fig:f4}
\end{figure*}

The base locomotion unit serves as the fundamental building block of the robot platform. Each unit consists of three linear actuators and two legs. Each actuator is connected at one end to a rigid octagonal frame and at the other end to a central leg connector. A single unit can move independently, and one complete walking cycle is illustrated in \textbf{Figure 4B}. At the neutral state, the unit is positioned at a slight tilt relative to the ground, with its legs and the front edge of the octagonal frame in contact with the surface. In Step 1, DEA 1 is actuated, driving the leg connector downward to increase friction between the leg and the ground. While DEA 1 remains actuated, DEA 2 is then activated, pushing against the leg connector; the resulting reactive force propels the octagonal frame forward, shifting the unit’s center of gravity and producing displacement. In Step 3, DEAs 1 and 2 return to their neutral state, and DEA 3 is actuated to lift and reposition the leg forward, preparing for the next walking cycle. The walking cycle is illustrated in \textbf{Movie S2}, and the analytical modeling of the robot is in \textbf{Supplementary Text}.

Signal frequency has orders of magnitudes influences on walking speed. The frequency response of walking speed was characterized for three robot configurations: a single locomotion unit (single-unit), two connected locomotion units (double-unit), and a quadruped robot composed of four units (quadruped bot). All robots were tethered, powered by a high-voltage benchtop supply (Trek 610E). Video footage of all robot configurations in operation is presented in \textbf{Movie S3}. Actuation sequences were generated through a custom switch board interfacing the actuators with the power supply. The actuation frequency was defined as the inverse of one complete walking cycle. The electric field magnitude was held constant at 40 $V/\mu m$. As shown in \textbf{Figure 4C}, the single-unit robot reached a peak speed of 114 $\text{cm/min}$ at 30 Hz, over 100 times faster than its speed at 1 Hz. The peak speed and resonant frequency decreased with larger assemblies: the double-unit achieved 71 $\text{cm/min}$ at 25 Hz, while the quadruped bot reached 22 $\text{cm/min}$ at 6 Hz. Frequency sweeps were terminated once each robot ceased to produce measurable displacement. Walking speed comparisons are provided in \textbf{Movie S4}.

The robot demonstrates long-term operation stability and payload capacity. A base locomotion unit is maintained operating for 3 hours at a resonant frequency at $40,V/\mu m$ and measured walking speed throughout the time. The walking speed remains constant, implying actuator long-term stability. The payload carrying capacity of the tethered robot was characterized to guide the design of untethered electronics. A resonant frequency of 6 Hz was selected for the quadruped robot. The robot’s speed was recorded under various payloads, as shown in \textbf{Figure 4E}. The tethered quadruped can carry over 200 g, exceeding 100\% of its body mass (183 g) and more than 700\% of the combined actuator mass (28 g for 12 actuators), while maintaining locomotion. The untethered electronics were designed to weigh between 100 and 200 g (137 g for components and 167 g for the assembled circuit). The circuit diagram is shown in \textbf{Figure S6}. The robot carrying electronics is shown in \textbf{Figure 4F}. However due to the power limitation of the compact high voltage convereters, the electronics cannot sufficiently drives the actuators to achieve locomotion. The system demonstrated here is therefore a prototype, and future work will address this power autonomy challenge.

The proposed robot is the first known DEA-powered quadruped capable of carrying payloads, demonstrating the potential of lightweight, quiet, and non-magnetic compliant actuators as alternatives to bulky, rigid counterparts. However, several key limitations remain to be addressed for achieving practical functionality. First, the robot’s open-loop walking trajectory is asymmetric, caused by actuator fabrication inconsistencies and the absence of feedback control. Second, the compact off-the-shelf high-voltage converters currently available (1 W) cannot deliver sufficient power to drive multiple scaled-up linear actuators, while higher-power converters are substantially heavier and exceed the robot’s payload capacity. Future work will require increasing the robot’s load-carrying capability, either by further scaling up the DEAs or by improving the overall robot architecture. Third, the walking cycle design, actuator configuration, and weight distribution of each locomotion unit can be further optimized to improve speed and reduce performance variation across trials.

\subsection*{Discussion}
In this work, we developed a multi-channel, multi-parameter lifetime testing robot for linear DEAs. Using this platform, we established a pipelined approach for DEA lifetime optimization, from sample testing to robotic application. We first scanned a parameter space defined by electric field, frequency, electrode material, and electrical connection filler using batch-fabricated, simplified linear DEAs. The optimal parameter combination was then used to scale the actuators in length and layer number, achieving greater force and displacement output. Finally, the scaled-up DEAs were integrated into a modular quadruped walking robot that demonstrated remarkable payload capacity and long-term stability.

Inspired by the growing adoption of self-driving labs (SDLs) in materials science and chemistry \cite{abolhasani2023rise, tom2024self}, this work represents the first step toward extending the SDL approach to device-level discovery and optimization, specifically in the domain of soft robotic actuators. While our system is not yet a fully autonomous SDL, it establishes the experimental and methodological foundation for one. We envision this pipeline as a direction-setting effort, guiding the community toward realizing closed-loop, device-level SDLs capable of accelerating actuator innovation. The remaining gaps, limitations, and our vision for a fully autonomous system are discussed in detail below.

\subsubsection*{Limitation}
The limitations of this work lie in hardware reliability, manual sample preparation, and limited data collection throughput. The testing robot is built largely from open-source electronics, which are generally reliable and cost-effective, but integration at the system level introduces reliability challenges. For instance, the opto-couplers experience overheating during extended operation, leading to occasional breakdowns before testing completes; cooling fans were added as a temporary solution. The use of high-voltage electronics also demands careful circuit design to prevent occasional voltage surges during system startup which can damage components. These reliability issues require periodic manual maintenance, reducing system autonomy and causing sample waste. Hardware design thus needs iterative refinement; there are no shortcuts in this process.

All DEA samples in this study were manually fabricated. Although 16 devices were made per batch, the process remains time-consuming. A full batch, including multilayer fabrication and electrical connection, takes roughly eight hours. This constraint limits both the complexity of the parameter space we can explore and the total data volume we can collect. Achieving a fully autonomous device-optimization system will require automated fabrication with combinatorial capabilities on ingredients.

Data acquisition (DAQ) throughput is another bottleneck. In this work, the DAQ system uses low-cost chips communicating with the Arduino via the I2C protocol, with data transmitted to a PC through USB. This architecture is reliable and sufficient for current needs, but scaling to a larger number of testing channels or higher sampling rates ($>$100 Hz per channel) will require advanced DAQ solutions such as FPGA-based interfaces \cite{prabhu2020elevating} to handle higher data throughput.

\begin{figure*}[!ht]
\centering
\includegraphics[width=\linewidth]{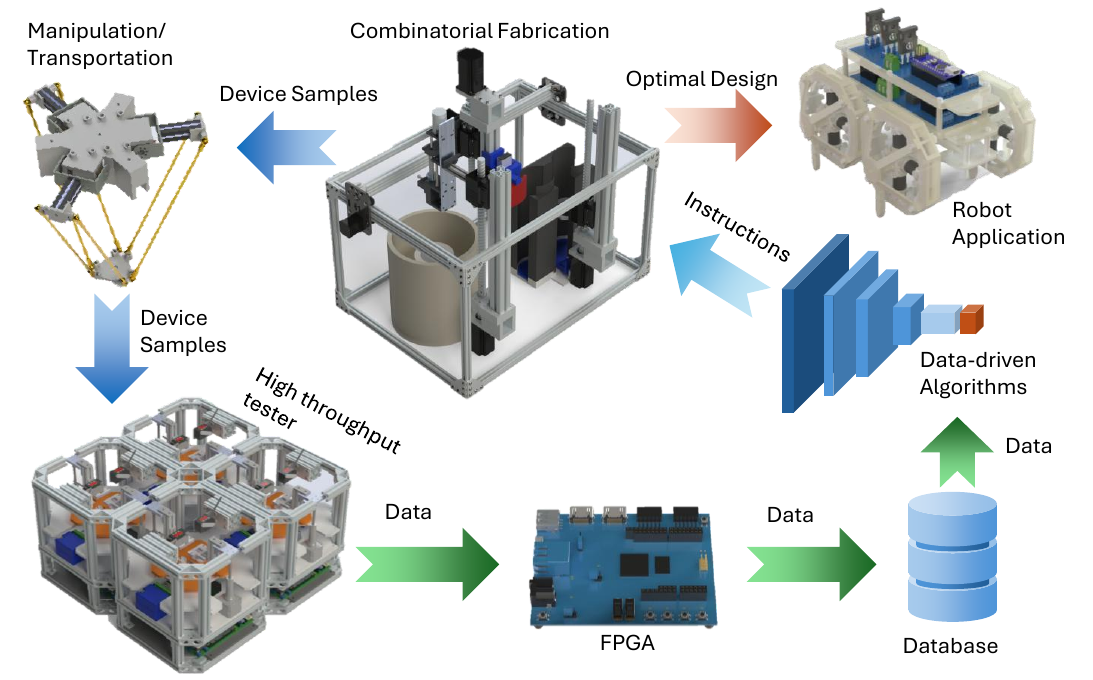}
\caption{\textbf{The vision of a generalizable device-level SDL for DEA design optimization}. A combinatorial fabrication robot (under development) produces device samples, which are transferred by a dexterous handling robot to the testing platform. The high-throughput testing robot conducts comprehensive lifetime measurements, while an FPGA-based node manages data collection and uploads data to a database. Data-driven algorithms are used to generate new fabrication parameters for subsequent iterations. The process converges toward an optimized material and device design, which is then transferred for robotic application.}
\label{fig:f5}
\end{figure*}

\subsubsection*{Vision}
The major challenge in extending SDLs from material-level to device-level discovery lies in shifting from single-variable, static measurements to multi-variable, time-domain data acquisition required for reliability testing under varied conditions. Using DEA optimization as a device-level example, degradation across materials and operating regimes is extremely challenging to be analytically or computationally modeled. Progress on optimization relies on large-scale experimental data and data-driven models for performance prediction and rapid screening of design choices. A fully autonomous device-level SDL must integrate combinatorial fabrication, sample handling, and high-throughput, multi-variable testing to close the loop.

 Device-level SDLs must vary both material composition and fabrication methods. Current systems typically tune one material through a fixed deposition route, whereas devices like DEAs combine electrodes and elastomers with distinct processes. For example, electrodes can be deposited by filter stamping \cite{duduta2016multilayer}, spraying \cite{peng2021stable}, or inkjet printing\cite{baechler2016inkjet}, while elastomers may be applied by spin coating \cite{lotz2010fabrication}, blade coating \cite{jiang2022long}, or 3D printing \cite{chortos20203d, tugui2023all}. Each method leads to different device behavior, requiring the fabrication platforms to adapt to diverse materials and processes. Device assembly and testing require complex and delicate manipulation. In DEAs, multilayer fabrication is followed by cutting, electrical connection, and placement for long-term testing, all still performed manually. Automation will require dexterous, benchtop-scale robotic platforms capable of programmable operations for packaging, loading, and transfer between fabrication and testing stages.

Application-oriented testing demands multi-channel, multi-variable setups. For DEAs, meaningful evaluation of force, displacement, and efficiency requires long-term operation under various waveforms, frequencies, and environmental conditions such as temperature \cite{albuquerque2021influence}, pressure \cite{li2021self}, or external load\cite{duduta2019realizing}. Battery cyclers provide a good example of a programmable, multi-channel device tester. However, unlike batteries, which undergo stationary testing, actuators and sensors need dynamic electromechanical measurements, requiring a higher degree of automation and procedure integration. Testing systems must therefore balance throughput with flexibility across applications.

 A device-level SDL should use collected data to iteratively fabricate and evaluate new devices, converging toward optimal design choices. The envisioned system should remain compact and lab-compatible, leveraging open-source electronics instead of industrial-scale testing facilities. Though development will require time and investment, such a system could greatly accelerate the transition from laboratory demonstration to practical application. DEAs serve here as an illustrative case, but the broader vision is a generalizable SDL framework for device-level discovery and optimization across soft robotics and related technologies.

\subsection*{Materials and Methods}
\subsubsection*{DEA materials and fabrication}
Linear DEAs are fabricated by first producing a multilayer DEA sheet, trimming excess elastomer to expose an edge for electrical connection, then rolling the sheet and encapsulating it between two rigid caps \textbf{(Figure S1)}. The cap assembly is designed to maximize the effective conductive area. An adhesive is used to attach an ultra-flex copper wire to a circular conductive metallic sheet (aluminum or copper), increasing the contact surface. A conductive filler material is applied to the exposed elastomer–electrode interface and serves as the colloidal layer between the elastomer and the metallic sheet. The multilayer fabrication procedures follow established methods \cite{zhao2018compact, duduta2016multilayer}. The caps are 3D printed using a Form 4 resin printer (Formlabs). Elastosil P7670 (Wacker Chemie AG) is selected as the elastomer, and CNT ink is prepared in-house using P3 CNT powder (Carbon Solutions, Inc.), following the published procedure \cite{hajiesmaili2021dielectric}. Ink consistency is verified by measuring optical transmittance and adjusting the density accordingly.

\subsubsection*{Conductive filler}
Three conductive filler materials are examined in this work: carbon grease (CG), eutectic gallium–indium liquid metal (LM) \cite{khan2014giant}, and a PDMS–carbon black mixture (CB) \cite{rosset2013flexible}. Carbon grease is an off-the-shelf product (MG Chemicals), while LM and CB are prepared in-house. The procedure for preparing LM is as follows: a 1:2 weight ratio of indium to eutectic gallium is heated to the liquid state in a 70,$^\circ$C water bath for 1 hour, then mixed on a hot plate at 60,$^\circ$C for 12 hours or until uniform. The procedure for preparing CB is as follows: the mixture is formulated using a 100:3 weight ratio of PDMS (part B) to carbon black powder as the conductive medium. Ingredients are mixed in a two-axis centrifugal mixer (FlackTek) at 2500 RPM for 2 minutes, with fractional additions of carbon black until the target concentration is reached. Multiple spin cycles ensure even particle distribution throughout the mixture.

\subsubsection*{Measurement of DEA properties}
Displacement of the DEAs is measured using a laser displacement sensor (Panasonic HGC-1030 series). A small piece of white paper is attached to the top surface of each DEA to provide a stable reflective surface. Blocked force is measured using FSG005WNPB force sensors (Honeywell). These sensors are mounted on linear actuators that automatically clamp onto the DEA samples to restrict motion. The clamping pressure is maintained using force-feedback from the sensors themselves, with a bias force of 0.6 N for DEA testing samples and 1 N for scaled-up actuators. Operation of the linear actuators is shown in \textbf{Movie S1}. Current and voltage feedback are obtained directly from the monitoring channels of the high-voltage converters (XP Power HRC series). Capacitance is measured using the impedance meter function of the Analog Discovery 3 (Digilent), with each sweep performed from 1 kHz to 1 MHz. Meaured capacitance prior and after long-term actuation for some DEA samples is ploted in \textbf{Figure S7}. Higher actuation voltages lead to greater capacitance degradation, while the influence of actuation frequency on degradation is less obvious.

\subsubsection*{Programmable high voltage power supply}
The dual-channel high-voltage power supply is built around programmable high-voltage converters. The circuit is modular and can accommodate either XP Power HRC-series 5W converters or HRL30-series 30W converters, depending on the power requirements of the DEAs (5W is sufficient for the Elastosil-based actuators used in this work). The circuit diagram is shown in \textbf{Figure S3}. This modular design enables straightforward scalability; additional converter channels can be added by connecting to the shared power line and linking their programming and feedback channels to the MCU. High-frequency square-wave actuation is generated using a dual opto-coupler configuration (HVM OR-100 series), where one opto-coupler manages charging, and the other manages discharging, operating in a complementary flip–flop manner so that only one is active at a time.

\subsubsection*{Walking robot}
The rigid framework components of the walking robot are printed using a Form 4 printer with standard clear SLA resin. All rigid parts are assembled using nylon or metallic screws. DEAs are mounted to the frame using double-sided tape. Sandpaper can be attached to the bottom of the legs to increase ground friction, and weight distribution can be adjusted in a modular fashion by attaching small magnets to the metallic screws.

Long-term speed tracking of the tethered quadruped robot is challenging due to cable length constraints and the lack of closed-loop control to maintain continuous walking. To assess long-term performance, we alternate between two measurement modes: allowing the robot to walk freely over a fixed distance to record ground speed, and blocking its motion with an obstacle while it continues actuating so that the actuators accumulate operating time. Repeating these two modes over a three-hour window provides an effective estimate of long-term speed trends while accommodating the limitations of a tethered setup.


\clearpage 
\bibliography{science_template} 
\bibliographystyle{sciencemag}


\section*{Acknowledgments}
The authors acknowledge prof. Pakpong Chirarattananon, prof. Jason Hattrick-Simpers from University of Toronto, and prof. Daniel van der Weide from University of Wisconsin-Madison for meaningful discussions.
\paragraph*{Funding:}
Natural Sciences and Engineering Resesarch Council of Canada - Discovery Grant RGPIN-2021-02791.
AFRL MidAtlantic Hub - Prototyping Award. 
University of Connecticut - SURF Award Summer 2025 (M. Francoeur).
Naval Institute for Undersea Vehicle Technologies - SEED 61. 
NASA Connecticut Graduate Fellowships (A. White, S. Sandhu, and V. Jimenez-Santiago).
\paragraph*{Author contributions:}
A.L. and M.D. initiated the research concept. A.L. built the testing robot. A.L., S.S., V.J-S., fabricated the DEAs. A.L. performed data collection and processing. A.W. designed the programmable power supply. A.Y. developed the walking cycle and analytical modeling for the walking robot. A.L. built and tested the walking robot. M.F. conducted material characterization experiments. V.R. helped with figure generation and vision development. C.T. made and tested electrical filler materials. A.L. wrote the paper. M.D. edited the paper.
\paragraph*{Competing interests:}
There are no competing interests to declare.
\paragraph*{Data and materials availability:}
All data needed to evaluate the conclusions in the paper are present in the paper and/or the Supplementary Materials.


\subsection*{Supplementary materials}
Supplementary Text\\
Figs. S1 to S7\\
Movies S1 to S4\\


\newpage


\renewcommand{\thefigure}{S\arabic{figure}}
\renewcommand{\thetable}{S\arabic{table}}
\renewcommand{\theequation}{S\arabic{equation}}
\renewcommand{\thepage}{S\arabic{page}}
\setcounter{figure}{0}
\setcounter{table}{0}
\setcounter{equation}{0}
\setcounter{page}{1} 


\begin{center}
\section*{Supplementary Materials for\\ \scititle}


Ang~(Leo)~Li$^{\ast}$,
Alexander~Yin,
Alexander~White, 
Sahib~Sandhu,
Matthew~Francoeur,\\
Victor~Jimenez-Santiago,
Van~Remenar,
Codrin~Tugui,
Mihai~Duduta$^{\ast}$\\
\small$^\ast$Corresponding author. Email: shffleo.li@mail.utoronto.ca , mihai.duduta@uconn.edu\\
\end{center}

\subsubsection*{This PDF file includes:}
Supplementary Text\\
Figures S1 to S7\\
Captions for Movies S1 to S4\\

\subsubsection*{Other Supplementary Materials for this manuscript:}
Movies S1 to S4\\

\newpage

\subsection*{Supplementary Text}
\subsubsection*{Analytical modeling of a base locomotion unit robot}
To gain a better understanding of how the a singular wheel moves, a simplified model was created to simulate the leg's motion and subsequently the amount of force it can generate due to the deformation of the attached DEAs. The single wheel has an outer octagon frame that has three DEAs attached to a central block connected to a leg. Activating the DEAs cause the central block and its attached leg to move. This motion causes the wheel to move as it shifts the center of gravity and exerts force through the leg that is touching the floor. The length of the leg and the angle at which it is attached to the frame will affect how the wheel is positioned with respect to the floor. The leg is designed to extend slightly outside the frame, causing the wheel to be positioned at a slight angle with respect to the floor. In this configuration, the wheel will only have two points of contact with the floor, the leg and a corner of the frame, shown in \textbf{Figure 4B}. As the DEAs are powered, the position of the leg will shift causing the body to change its angle with the floor. The center block's location can be defined using its relation to the frame. Its height compared to the bottom of the frame is defined at \(h_c\) and its horizontal distance from the pivot corner is defined as \(w_c\) both in centimeters. Its location will change depending on which DEA is active as indicated by \(e_n\) which defines the linear elongation for the given DEA, \(n\). To simplify the simulation, all DEAs are assumed to be ideal such that elongation, \(e_n\) and force \(F_n\) are the maximum recorded values from testing. Additionally, it is assumed that they are not antagonistic to one another even though they are all attached to the center block because they are flexible and the elongation is small. This results in equations 1 and 2.

\begin{equation}
    h_c = \frac{h}{2} - e_1 + e_3
\end{equation}

\begin{equation}
    w_c = \frac{b}{2} + e_2
\end{equation}

Using the central block's position and the leg's configuration, further information about how the leg is positioned with respect to the frame can be calculated. The length of the leg \(l\) and the attached angle \(\theta_l\) can be used to calculate the vertical length between the end of the leg and the bottom of the frame, \(\delta_h\) shown in equation 3. This information can then be used to calculate the length of the leg that is outside the frame, \(\delta_l\), shown in equation 4. This is then used to find the horizontal distance between the pivot point and the point the leg crosses over the base, \(\delta_w\) shown in equation 5.

\begin{equation}
    \delta_h = l\cos(\theta_l) - h_c
\end{equation}

\begin{equation}
    \delta_l = \delta_h\sec(\theta_l)
\end{equation}

\begin{equation}
    \delta_w = w_c + (l - \delta_l)\sin(\theta_l)
\end{equation}

With the additional information, the distance between the two contact points on the floor, \(d\), can be calculated, as shown in figure 5. This can then be used to find the tilt the body is compared to the ground denoted as \(\theta_b\) and shown in equation 7.

\begin{equation}
    d = \sqrt{\delta_l^2 + \delta_w^2 - 2\delta_l\delta_w\cos{(\theta_l + 90)}}
\end{equation}

\begin{equation}
  \theta_b = \arcsin\bigg(\frac{\delta_l\sin(\theta_l + 90)}{d}\bigg)
\end{equation}

The body angle is then used to understand the internal forces that are created by the attached DEAs. The forces induced by the DEAs change depending on the posture of the frame. For further simplification, the forces are split into its vertical \(F_y\) and horizontal \(F_x\) components shown in equations 8 and 9.

\begin{equation}
    F_x = (-F_1+F_3)*\sin(\theta_b) - F_2\cos(\theta_b)
\end{equation}

\begin{equation}
    F_y = (-F_1+F_3)*\cos(\theta_b) - F_2\cos(\theta_b)
\end{equation}

With the simulation, different leg configurations, DEA specifications, and pulse timings can be explored to further improve the wheel's performance. Currently, a simple 3 step open loop control is used for a single wheel shown in the \textbf{movie s1}. More complex controls involving different pulse timings such as a descending ramp to turn off the DEA could prove useful to the wheels performance. 

\newpage

\begin{figure} 
	\centering
	\includegraphics[width=0.9\textwidth]{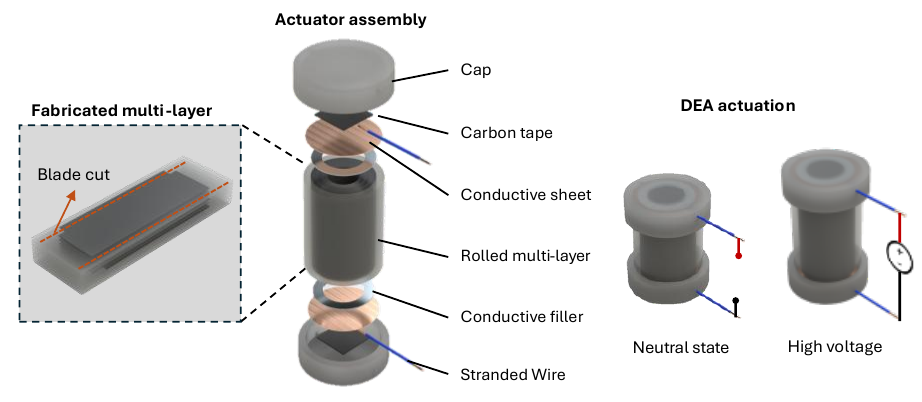} 

	\caption{\textbf{Linear actuator assembly. }
		The multilayer actuator is fabricated using standard procedures well-documented in literature \cite{zhao2018compact, duduta2016multilayer}. After trimming, the multilayers are rolled and placed between two end caps. Carbon tape provides an adhesive interface for conductors, connecting stranded copper wires and conductive sheets. The wires serve as external electrical leads, while the conductive sheets extend the effective contact area with the DEA. A conductive filler layer is applied between the conductive sheet and the multilayer cross section; this layer is critical to the device’s cycle-life performance and is a key focus of material selection in this work. The actuator expands linearly along its length when an electric field is applied across its terminals.
		}
	\label{fig:s1} 
\end{figure}

\begin{figure} 
	\centering
	\includegraphics[width=0.9\textwidth]{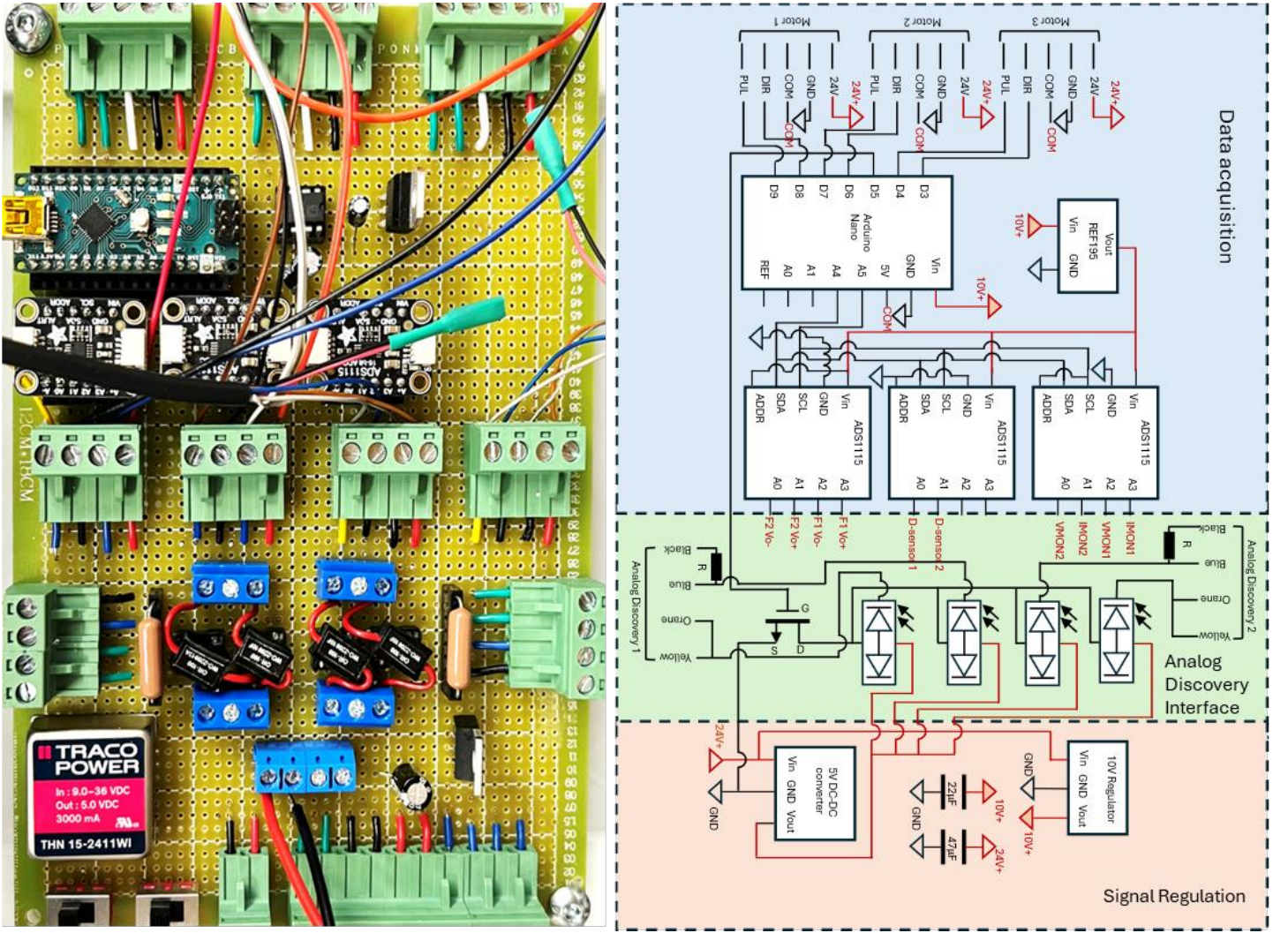} 

	\caption{\textbf{Data acquisition circuit for the testing robot.}
	(Left) the photo of the circuit board. (Right) The circuit topology.
		}
	\label{fig:s2} 
\end{figure}

\begin{figure} 
	\centering
	\includegraphics[width=0.9\textwidth]{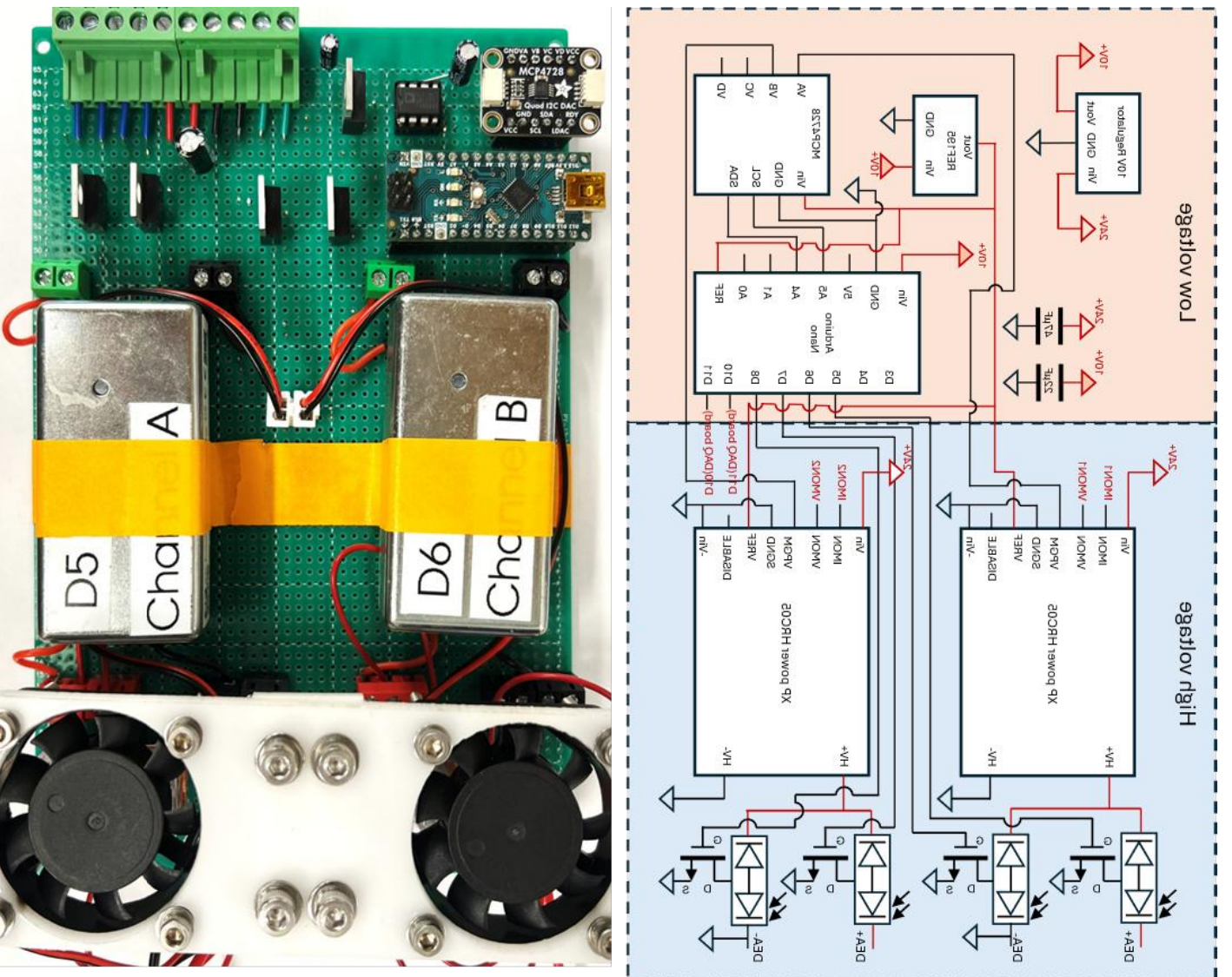} 

	\caption{\textbf{Programmable power supply circuit for the testing robot.}
	 (Left) the photo of the circuit board. (Right) The circuit topology (cooling fans are not included).
		}
	\label{fig:s3} 
\end{figure}

\begin{figure} 
	\centering
	\includegraphics[width=0.9\textwidth]{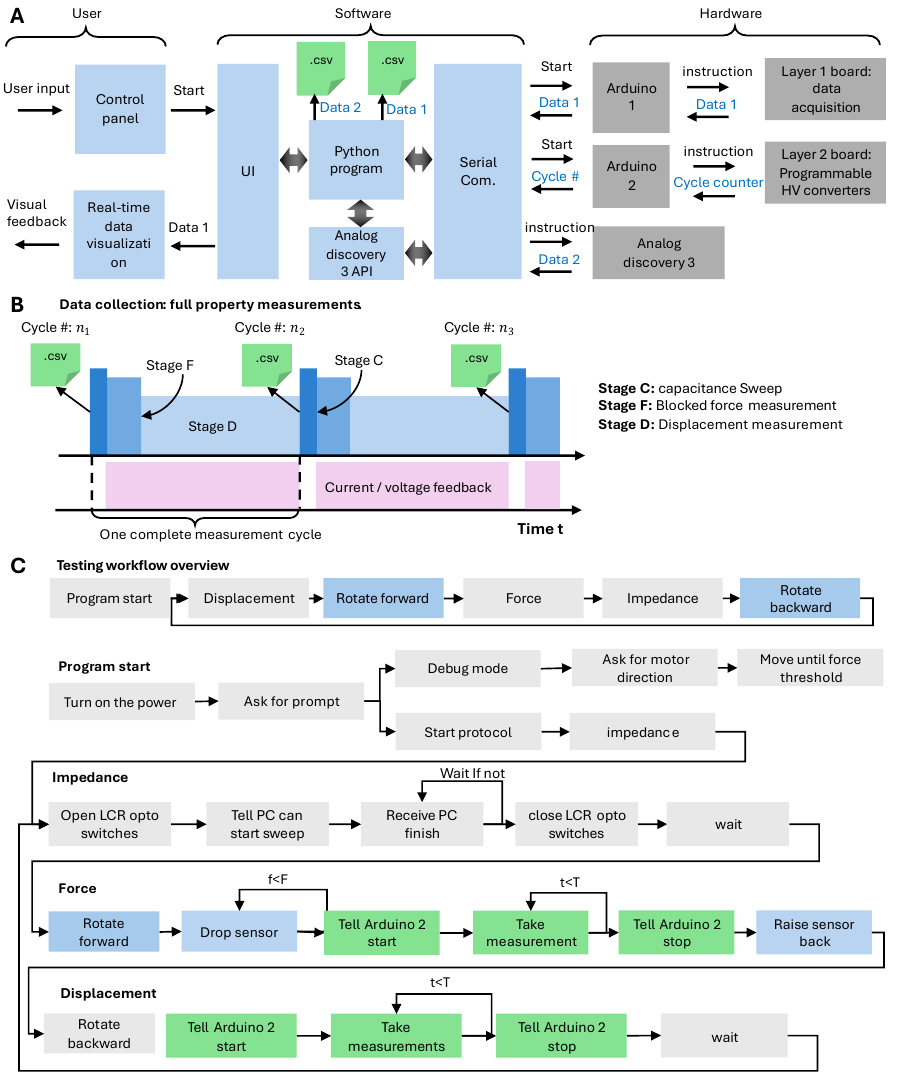} 

	\caption{\textbf{Testing robot architecture and workflow.}
    (A) System architecture from user interface to hardware. (B) Data collection sequence for full electromechanical measurements. The robot periodically switches measurement modes to capture complete datasets; however, most data in this work were collected using simplified protocols measuring either force or displacement. (C) Workflow of Arduino 1 (top-layer control and DAQ board). Motor actions are shown in blue, and Arduino 2 (high-voltage board) actions in green.
		}
	\label{fig:s4} 
\end{figure}

\begin{figure} 
	\centering
	\includegraphics[width=0.9\textwidth]{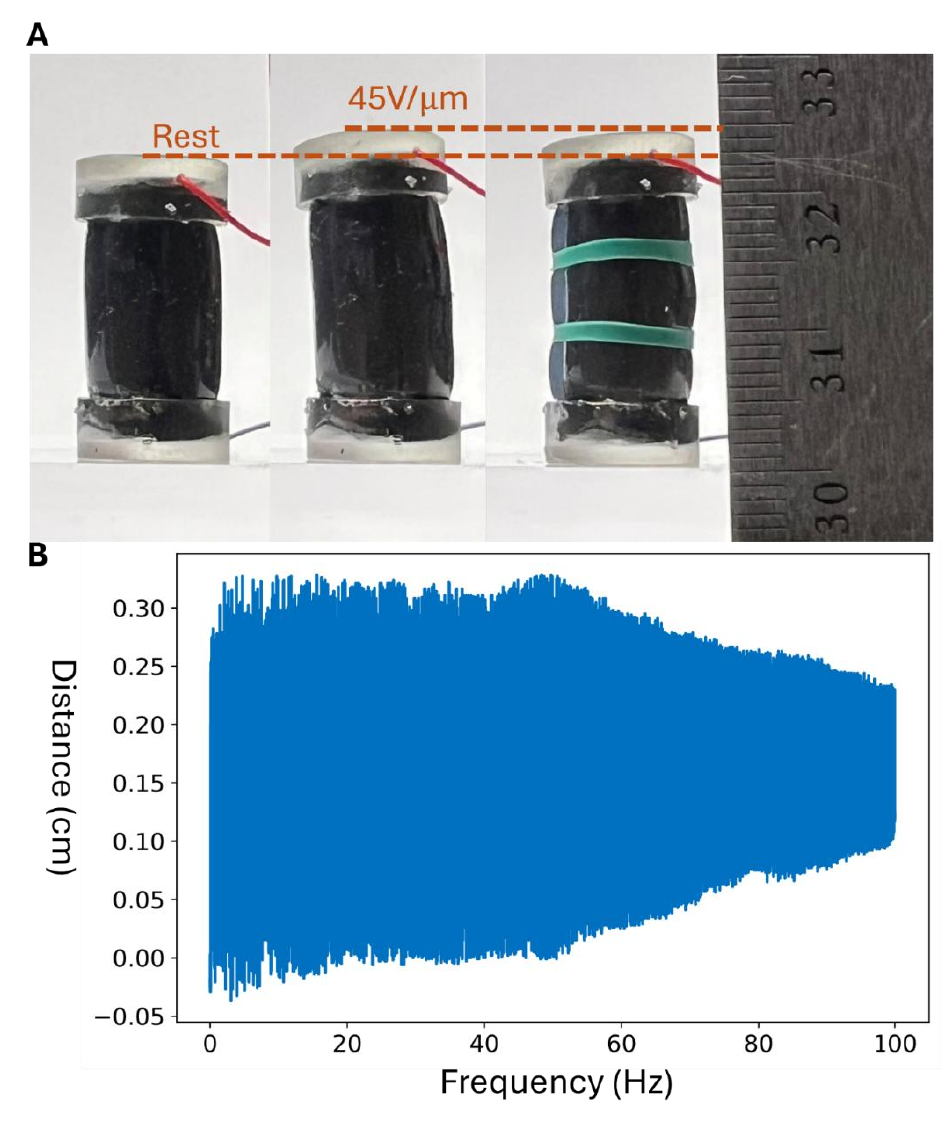} 

	\caption{\textbf{Displacement performance of the scaled-up actuator.}
	(A) Comparison of DEA actuators with and without ribbon reinforcement at 42 $V/\mu m$. No significant difference is observed between the two configurations, both achieving about 2 mm displacement (8.7\% axial strain). (B) Displacement versus frequency response of the scaled-up actuator at 45 $V/\mu m$, obtained using a sweep signal ranging from 1 Hz to 100 Hz over 100 seconds.
		}
	\label{fig:s5} 
\end{figure}

\begin{figure} 
	\centering
	\includegraphics[width=0.9\textwidth]{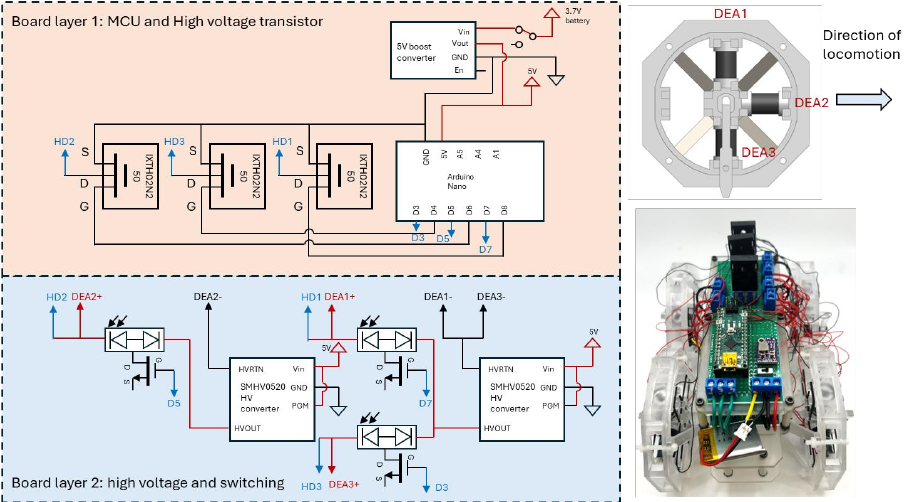} 

	\caption{\textbf{Circuit diagram for untethered quadruped robot power electronics.}
	  The top layer of the board consists of MCU, battery voltage converter, and high voltage MOSFETs for discharge. The bottom layer of the board consists of two high voltage converters (HVM  SMHV0520) and 3 high voltage opto-couplers (HVM OR-100).
		}
	\label{fig:s6} 
\end{figure}

\begin{figure} 
	\centering
	\includegraphics[width=0.9\textwidth]{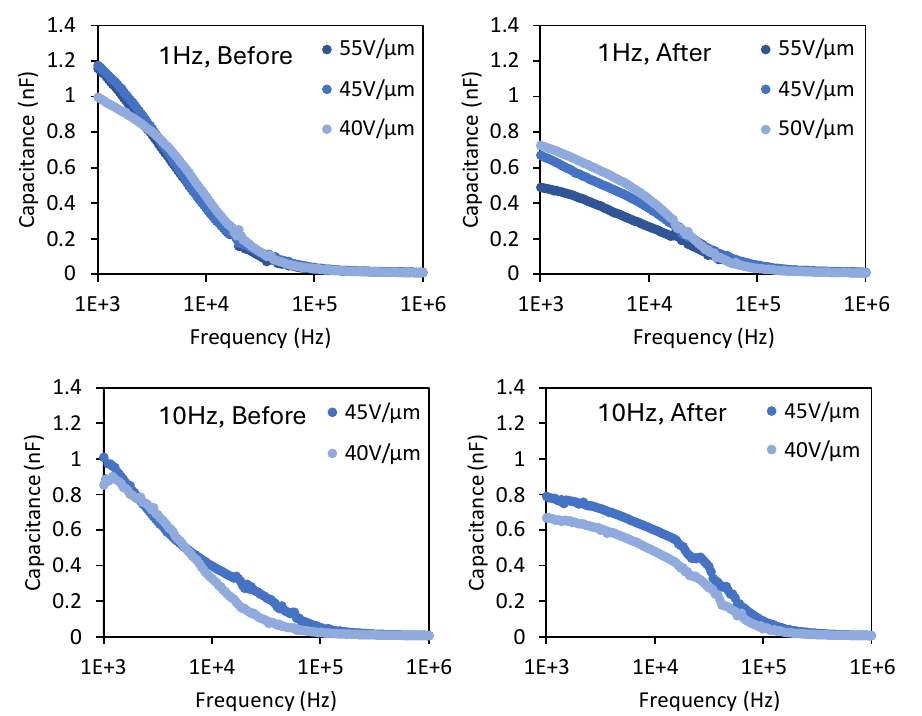} 

	\caption{\textbf{Capacitance–frequency sweep measurements of device samples under different electric fields and actuation frequencies.}
	  Measurements were performed using Analog Discovery 3 modules mounted on the testing robot. The left plots show capacitance before testing, and the right plots show capacitance after testing. Higher actuation voltages lead to greater capacitance degradation, while the influence of actuation frequency on degradation is less pronounced.
		}
	\label{fig:s7} 
\end{figure}


\clearpage 

\paragraph{Supporting video files are available upon request.}
\paragraph{Caption for Movie S1.}
\textbf{Motor stage of the testing robot in action}
This video shows top and side views of the testing robot switching from displacement measurement to force measurement and then back to displacement. The rotational stepper motor moves the actuator between the two measurement positions, and the linear stepper carrying the force sensor clamps onto the actuator using force feedback. When switching back, both the linear and rotational motors return to their original positions.

\paragraph{Caption for Movie S2.}
\textbf{Animation of the walking cycle of a base locomotion robot}
The analytical modeling of this system is in the Supplementary Text.

\paragraph{Caption for Movie S3.}
\textbf{Operation of different robot configurations.}
The robot configurations include (i) base locomotion unit, (ii) double unit, (iii) quadruped walking robot.

\paragraph{Caption for Movie S4.}
\textbf{Robot operation under different frequencies and payload.}
The walking speeds of a base locomotion unit, a double unit, and a quadruped robot are compared at 1 Hz and at their respective resonant frequencies. A quadruped robot carrying payload operates at its resonant frequency of 6 Hz.



\end{document}